\documentclass{article} 
\usepackage{nips12submit_e,times}
\usepackage{graphicx}
\usepackage{amssymb}
\usepackage{epstopdf}

\newcommand{\la}{\lambda}
\newcommand{\La}{\Lambda}

\newcommand{\om}{\omega}

\title{Learning Stable Group Invariant Representations with Convolutional Networks}

\author{
Joan Bruna, Arthur Szlam and Yann LeCun\\
Courant Institute \\
New York University\\
New Nork, NY, 10013 \\
\texttt{\{bruna,lecun\}@cims.nyu.edu} 
}

\nipsfinalcopy 

\begin{document}

\maketitle

%

\section{Introduction}

Many signal categories in vision and auditory 
problems are invariant to the action of transformation
groups, such as translations, rotations or frequency
transpositions. 
This property motivates the study of signal representations
which are also invariant to the action of these
transformation groups. For instance, translation 
invariance can be achieved with a registration 
or with auto-correlation measures. 

Transformation groups are in fact low-dimensional
manifolds, and therefore mere group invariance 
is in general not enough to efficiently describe 
signal classes. Indeed, signals may be perturbed with 
additive noise and also with geometrical deformations, 
so one can then ask for invariant representations which
are stable to these perturbations. Scattering 
convolutional networks \cite{scatt_pami} construct
locally translation invariant signal representations,
 with additive and geometrical
stability, by cascading
complex wavelet modulus operators with a 
lowpass smoothing kernel. 
By defining wavelet decompositions
on any locally compact Lie Group, scattering
operators can be generalized and cascaded 
to provide local invariance with respect to more 
general transformation groups \cite{scatt_steph, scatt_laurent}.
Although such transformation groups are present
across many recognition problems, they 
require prior information which sometimes 
cannot be assumed. 

Convolutional networks \cite{convnet_yann} 
cascade filter banks with
point-wise nonlinearities and local pooling operators. 
By remapping the output of each layer with the input
of the following one, the trainable filters implement 
convolution operators. 
We show that the invariance properties built
by deep convolutional networks can be 
cast as a form of stable group invariance. The network
wiring architecture determines the invariance group, 
while the trainable filter coefficients characterize
the group action. 

Deep convolutional architectures cascade several layers of convolutions, 
non-linearities and pooling. 
These architectures have the capacity to generate 
local invariance to the action of more general groups. 
Under appropriate conditions, these groups can 
be factorized as products of smaller groups.
Each of these factors can then be associated
with a subset of consecutive layers of the convolutional
network.
In these conditions, 
the invariance properties of the final representation can be 
studied from the group structure generated by each 
layer. 


%

\section{Problem statement}

\subsection{Stable Group Invariance}
A transformation group $G$ acts 
on the input space $\mathcal{X}$ 
(assumed to 
be a Hilbert space) with a 
linear group action 
$(g,x) \mapsto g.x \in \mathcal{X}$,
which is compatible with the group operation.

A signal representation $\Phi: \mathcal{X} \longrightarrow \mathcal{Z}$ 
is invariant to the action of $G$ if 
$\forall \, g \in G\,,x \in \mathcal{X}~,~ \Phi( g.x ) = \Phi(x) ~.$
However, mere group invariance is in general too weak, 
due to the presence of a much larger, high dimensional
variability which does not belong to the low-dimensional
group.
It is then necessary to incorporate the notion of 
 outer ``deformations" with another group action
 $\varphi: H \times \mathcal{X} \longrightarrow \mathcal{X}~,$
where $H$ is a larger group containing $G$.
The geometric stability can be stated 
with a Lipschitz continuity property
\begin{equation}
\label{lipseq}
\| \Phi( \varphi(h, x)) - \Phi( x) \| \leq C \,\|x \| \,k(h, G)~,
\end{equation}
where $k(h,G)$ measures the ``distance" from $h$ to
the invariance group $G$.
For instance, when $G$ is the translation group of $\mathbb{R}^d$
and $H \supset G$ is the group of $C^2$ diffeomorphisms of $\mathbb{R}^d$,
then $\varphi(h,x)=x \circ h$ and one can select as distance the elastic
deformation metric $k(h, G) := | \nabla \tau |_\infty + |H \tau|_\infty$, 
where $\tau(u)=h(u)-u$ \cite{scatt_steph}.

 Even though the group invariance formalism 
 describes global invariance properties of the representation, 
 it also provides a valid and useful framework to study local 
 invariance properties. Indeed, if one replaces (\ref{lipseq}) by
\begin{equation}
\label{loc_invariance_eq}
\| \Phi( \varphi(h, x)) - \Phi( x) \| \leq C \,\|x \| \,( \| h_{G} \|_G + k(h, G) )~,
\end{equation}
 where $h_G$ is a projection of $h$ to $G$ and 
 $\| g \|_G$ is a metric on $G$ measuring 
 the amount of transformation being applied, then
 the local invariance is expressed by 
 adjusting the proportionality between the two metrics.
 

\subsection{Convolutional Networks}
A generic convolutional network defined on a
space $\mathcal{X}=L^2(\Omega_0)$ of square-integrable 
signals starts with a filter bank $\{ \psi_\la \}_{\la \in \La_1}$, $\psi_\la \in L^1(\Omega_0) \, \forall \la$, 
which for each input $x(u) \in \mathcal{X}$ produces the collection
$$z^{(1)}(u, \la)  = x \star \psi_\la (u) = \int x(u-v) \psi_\la(v) dv~,~u \in \Omega_0\, , \la \in \La_1~.$$
If the filter bank defines a stable, invertible frame, then 
there exist two constants $a, A > 0$ such that
$$\forall x~, a \| x \| \leq \| z^{(1)} \| \leq A \| x \|~,$$
where $\| z^{(1)} \|^2 = \sum_{\la \in \Lambda_1} \| z^{(1)}(\cdot, \la) \|^2$.
By defining $\Omega_1 = \Omega_0 \times \La_1$, 
the first layer of the network can be written as the 
linear mapping
\begin{eqnarray*}
F_1:  L^2(\Omega_0) &\longrightarrow& L^2(\Omega_1) \\
 x(u) &\longmapsto& z^{(1)}(u,\la)~.
\end{eqnarray*}
$z^{(1)}$ is then transformed with a 
point-wise nonlinear operator $M: L^2(\Omega) \rightarrow L^2(\Omega)$ 
which is usually non-expansive, meaning that
$\| Mz \| \leq \| z \|$. 
Finally, a local pooling operator $P$  
can be defined as any 
linear or nonlinear operator
$$P: L^2(\Omega) \longrightarrow L^2(\widetilde{\Omega})$$
which reduces the resolution of the signal
along one or more coordinates and
which avoids ``aliasing". 
If $\Omega=\Omega_0 \times \La_1, É, \times \La_k$ 
and $(2^{J_0},É,2^{J_k})$ denote the loss 
of resolution along each coordinate,  
it results that 
$\widetilde{\Omega} = \widetilde{\Omega_0} \times \widetilde{\La_1}, É, \times \widetilde{\La_k}$, 
with $| \widetilde{\Omega_0}| =  2^{-\alpha J_0}|{\Omega_0}|$, 
$| \widetilde{\La_i}| =  2^{-\alpha J_i}|{\La_i}|$, where $\alpha$ 
is an oversampling factor.
Linear pooling operators are implemented as
lowpass filters $\phi_J(u,\la_1,É,\la_m)$ followed
by a downsampling. 

Then, a $k$-layer convolutional network is 
a cascade 
\begin{equation}
\label{convnet}
L^2(\Omega_0) \stackrel{M \circ F_1}{\longrightarrow} L^2(\Omega_1) \stackrel{P_1}{\longrightarrow} L^2(\widetilde{\Omega_1}) \stackrel{M \circ F_2}{\longrightarrow} L^2(\Omega_2) \,\cdots \, \stackrel{P_k}{\longrightarrow} L^2(\widetilde{\Omega_k}) ~,
\end{equation}
which produces successively $z^{(1)}, z^{(2)},\dots,z^{(k)}$.

The filter banks $(F_i)_{i \leq k}$, together with the pooling operators $(P_i)_{i \leq k}$, 
progressively transform the signal domain; filter bank steps 
lift the domain of definition by adding new coordinates, whereas pooling steps
reduce the resolution along certain coordinates. 

\section{Invariance Properties of Convolutional Networks}




\subsection{The case of one-parameter transformation groups}
Let us start by assuming the simplest form of 
variability produced by a transformation group. 
A \emph{one-parameter transformation group} 
is a family $\{U_t\}_{t \in \mathbb{R}}$ 
of unitary linear operators of $L^2(\Omega)$ such 
that (i) $t \mapsto U_t$ it is strongly continuous: $\lim_{t \to t_0} U_t z = U_{t_0} z$ for every $z \in L^2(\Omega)$, 
and (ii) $U_{t+s} = U_t U_s$.
One parameter transformation groups are thus homeomorphic
to $\mathbb{R}$ (with the addition as group operation), and define 
an action which is continuous in the group variable.
Uni-dimensional translations $U_t x(u) = x(u-tv_0)$, 
frequency transpositions $U_t x = \mathcal{F}^{-1}(\mathcal{F}x(\om-t \om_0))$ 
(where $\mathcal{F}$, $\mathcal{F}^{-1}$ are respectively the forward 
and inverse Fourier transform) 
or unitary dilations $U_t x(u) = 2^{-t/2} x(2^{-t}u)$ are examples
of one-parameter transformation groups.

One-parameter transformation 
groups are particularly simple to study
thanks to Stone's theorem \cite{stone}, 
which states that unitary one-parameter 
transformation groups are uniquely generated by 
a complex exponential of a self-adjoint operator:
$$U_t = e^{i t A}~,~t \in \mathbb{R}~.$$
Here, the complex exponential of a self-adjoint operator
should be interpreted in terms of its spectra.
In the finite dimensional case (when $\Omega$ is discrete), 
this means that there exists
an orthogonal transform $O$ such that
if $\hat{z}(\om) = O z$, then
\begin{equation}
\label{stone_complex}
\forall\, z~,~ U_t  z = O^{-1} \mbox{diag}(e^{i t \om}. \hat{z}(\om)) ~.
\end{equation}
In other words, the group action can be 
expressed as a linear phase change in the basis which
diagonalizes the unique self-adjoint operator $A$
given by Stone's theorem.
In the particular case of translations, the
change of basis $O$ is given by the Fourier
transform.
As a result, one can obtain a representation which is 
invariant to the action of $\{U_t\}_t$ with a 
single layer of a neural network: 
a linear decomposition which expresses the data 
in the basis given by $O$ followed by a point-wise
complex modulus. In the case of the translation 
group, this corresponds to taking the modulus
of the Fourier transform.

\subsection{Presence of deformations}

Stone's theorem provides a recipe for 
global group invariance
for strongly continuous group actions. 
Without noise nor deformations, an invariant
representation can be obtained by 
taking complex moduli in a basis which 
diagonalizes the group action, 
which can be implemented in a 
shallow $1$-layer architecture.
However, the underlying low-dimensional 
assumption is rarely satisfied, due
to the presence of more complex forms
of variability.

This complex variability can be modeled as follows.
If $O$ is the basis which diagonalizes a 
given one-parameter group, then the 
group action is expressed in the 
basis $\mathcal{F}^{-1} O$ 
as the translation operator $T_s z(u) = z(u-s)$. 
Whereas the group action consists in 
rigid translations on this basis, by
analogy a \emph{deformation} 
is defined as a non-rigid warping in this
domain: $L_\tau z(u) = z(u - \tau(u))$, 
where $\tau$ is a displacement field
along the indexes of the decomposition.

The amount of deformation can be measured
with the regularity of $\tau(u)$, which controls
how distant the warping is from being a rigid
translation and hence an element of the group.
This suggests that, in order to obtain stability
to deformations,
rather than looking for eigenvectors
of the infinitesimal group action, one should
look for linear measurements which are well
localized in the domain where deformations occur, 
and which nearly diagonalize the group action.
In particular, these measurements can be 
implemented with convolutions using compactly
supported filters, such as in convolutional networks.

Let $z^{(n)}(u,\la_1,\dots,\la_n)$ be an intermediate representation
in a convolutional network, 
and whose first layer is fully connected.
 Suppose that $G$ is 
a group acting on $z$ via
\begin{equation}
\label{groupsimple_action}
g.z^{(n)}(u,\la_1,\dots,\la_n) = z^{(n)}(u,\la_1+\eta(g),\la_2,\dots,\la_n)~,
\end{equation}
where $\eta : G \to \La_1$. 
This corresponds to the idealized case 
where the transformation only modifies one 
component of the representation.
A local pooling operator along the 
variable $\la_1$, at a certain scale $2^J$,
attenuates the transformation by $g$ as soon as
 $|\eta(g)| \ll 2^J$. It 
 thus produces local 
invariance with respect to the action of $G$.

\subsection{Group Factorization with Deep Networks}

Deep convolutional networks have the capacity
to learn complex relationships of the data and 
to build invariance with respect to a large
family of transformations. 
These properties can be partly explained 
in terms of a factorization of the invariance
groups performed successively.

Whereas pooling operators efficiently produce 
stable local invariance, convolution operators
preserve the invariance generated by previous 
layers. Indeed, suppose $z^{(n)}(u, \la_1)$ is an 
intermediate representation in a convolutional
network, and that $G$ acts on $z^{(n)}$ 
via $g.z^{(n)}(u,\la_1) = z^{(n)}(f(g,u), \la_1+\eta(g))$.
It follows that if the next layer is constructed as
$$z^{(n+1)}(u,\la_1,\la_2) := z^{(n)}(u,\cdot) \star \psi_{\la_2} (\la_1)~,$$
then $G$ acts on $z^{(n+1)}$ via
$g.z^{(n+1)}(u,\la_1,\la_2) = z^{(n)}(f(g,u), \la_1+\eta(g), \la_2)$, 
since convolutions commute with the group
action, which by construction is expressed as 
a translation in the coefficients $\la_1$. The new
coordinates $\la_2$ are thus unaffected by the action of $G$.

As a consequence, this property enables a 
systematic procedure to generate invariance
to groups of the form $G= G_1 \rtimes G_2 \rtimes \dots \rtimes G_s$, 
where $H_1 \rtimes H_2$ is the \emph{semidirect} product of groups. 
In this decomposition, each factor $G_i$ is associated
with a range of convolutional layers, along the coordinates
where the action of $G_i$ is perceived.
%

\section{Perspectives}


The connections between group invariance
and deep convolutional networks offer 
an interpretation of their efficiency on
several recognition tasks.
In particular, they might 
explain why the weight sharing induced by 
convolutions is a valid regularization 
method in presence of group variability.

More concretely, we shall also
concentrate on the following aspects:

\begin{itemize}
\item \emph{Group Discovery}. One might 
ask for the group of transformations which 
best explains the variability observed in 
a given dataset $\{x_i\}$. In the case where no geometric
deformations are present, one can start 
by learning the (complex) eigenvectors of the 
group action:
$$U^*= \mbox{arg}\min_{U^T U = {\bf 1}} \mbox{var}( |Ux_i| )~.$$
When the data corresponds to a uniform measure on 
the group, then this decomposition can be obtained 
from the diagonalization of the covariance operator $\Sigma= E(x^T x)$.
In that case, the real eigenvectors of $\Sigma$ are grouped into pairs
of vectors with identical eigenvalue, which then define the complex
decomposition diagonalizing the group action.

In presence of deformations, the global invariance
is replaced by a measure of local invariance. This problem 
is closely related to the sparse coding with slowness from \cite{cadieu}.

\item \emph{Structured Convolutional Networks}. 
Groups offer a powerful framework to incorporate 
structure into the families of filters, similarly
is in \cite{artur}. On the one hand, 
one can enforce global properties of the group by 
defining the convolutions accordingly. For instance, 
by wrapping the domain of the convolution, one is
enforcing a periodic group to emerge. 
On the other hand, one could further regularize the 
learning by enforcing a group structure within a filter 
bank. For instance, one could ask a certain filter
bank $F=\{h_1,\dots, h_n\}$ to have the form $F=\{R_\theta h_0\}_{\theta}$,
where $R_\theta$ is a rotation with an angle $\theta$. 

\end{itemize}

%
%

\end{document}